%% file: main.tex
\documentclass[acmtog]{acmart}

\usepackage{multirow}
\usepackage{amsmath}
\usepackage{makecell}
\input{preamble}

\input{definitions}
\usepackage{siunitx}
\usepackage{algorithm}
\usepackage{pifont}        %
\usepackage{enumitem}

\definecolor{heatYellow}{HTML}{FFFFB2}   %
\definecolor{heatOrange}{HTML}{FFD9B2}   %
\definecolor{heatRed}{HTML}{FFB2B2}   %
\definecolor{Bred}{HTML}{b5341e}   
\definecolor{Fgreen}{HTML}{309c59}

\newcommand{\cmark}{\textcolor{Fgreen}{\ding{52}}} %
\newcommand{\xmark}{\textcolor{Bred}{\ding{56}}}

\definecolor{myPink}{HTML}{E91E63}  %
\definecolor{black}{HTML}{000000}  %
\newcommand{\newtext}[1]{\textcolor{black}{#1}}

\begin{document}
\copyrightyear{2025}
\acmYear{2025}
\acmConference[SA Conference Papers '25]{SIGGRAPH Asia 2025 Conference Papers}{December 15--18, 2025}{Hong Kong, Hong Kong}
\acmBooktitle{SIGGRAPH Asia 2025 Conference Papers (SA Conference Papers '25), December 15--18, 2025, Hong Kong, Hong Kong}\acmDOI{10.1145/3757377.3763862}
\acmISBN{979-8-4007-2137-3/2025/12}

\title{PhySIC: Physically Plausible 3D Human-Scene Interaction and Contact from a Single Image}

\author{Pradyumna Yalandur Muralidhar}
\authornote{Equal contribution. \quad \textsuperscript{\dag}Corresponding Author.}
\affiliation{
\institution{University of Tübingen, Zuse School ELIZA }
\country{Germany}
}

\author{Yuxuan Xue}
\authornotemark[1]
\authornotemark[2]
\affiliation{
\institution{University of Tübingen, Tübingen AI Center}
\country{Germany}
}

\author{Xianghui Xie}
\affiliation{
\institution{University of Tübingen, Tübingen AI Center, MPI for Informatics, SIC}
\country{Germany}
}

\author{Margaret Kostyrko}
\affiliation{
\institution{University of Tübingen, Tübingen AI Center}
\country{Germany}
}

\author{Gerard Pons-Moll}
\affiliation{
\institution{University of Tübingen, Tübingen AI Center, MPI for Informatics, SIC}
\country{Germany}
}

\begin{abstract}
Reconstructing metrically accurate humans and their surrounding scenes from a single image is crucial for virtual reality, robotics, and comprehensive 3D scene understanding. 
However, existing methods struggle with depth ambiguity, occlusions, and physically inconsistent contacts.
To address these challenges, we introduce \textbf{PhySIC}, a unified framework for physically plausible Human–Scene Interaction and Contact reconstruction. PhySIC recovers metrically consistent SMPL-X human meshes, dense scene surfaces, and vertex-level contact maps within a shared coordinate frame, all from a single RGB image. 
Starting from coarse monocular depth and parametric body estimates, PhySIC performs occlusion-aware inpainting, fuses visible depth with unscaled geometry for a robust initial metric scene scaffold, and synthesizes missing support surfaces like floors. A confidence-weighted optimization subsequently refines body pose, camera parameters, and global scale by jointly enforcing depth alignment, contact priors, interpenetration avoidance, and 2D reprojection consistency. 
Explicit occlusion masking safeguards invisible body regions against implausible configurations. 
PhySIC is highly efficient, requiring only 9 seconds for a joint human-scene optimization and less than 27 seconds for end-to-end reconstruction process. 
Moreover, the framework naturally handles multiple humans, enabling reconstruction of diverse human scene interactions. 
Empirically, PhySIC substantially outperforms single-image baselines, reducing mean per-vertex scene error from 641 mm to 227 mm, halving the pose-aligned mean per-joint position error (PA-MPJPE) to 42 mm, and improving contact F1-score from 0.09 to 0.51. 
Qualitative results demonstrate that PhySIC yields realistic foot-floor interactions, natural seating postures, and plausible reconstructions of heavily occluded furniture. 
By converting a single image into a physically plausible 3D human-scene pair, PhySIC advances accessible and scalable 3D scene understanding. 
Our implementation is publicly available at \url{https://yuxuan-xue.com/physic}.

\end{abstract}



\begin{teaserfigure}
\begin{flushleft}
    \centering
    \includegraphics[width=\linewidth]{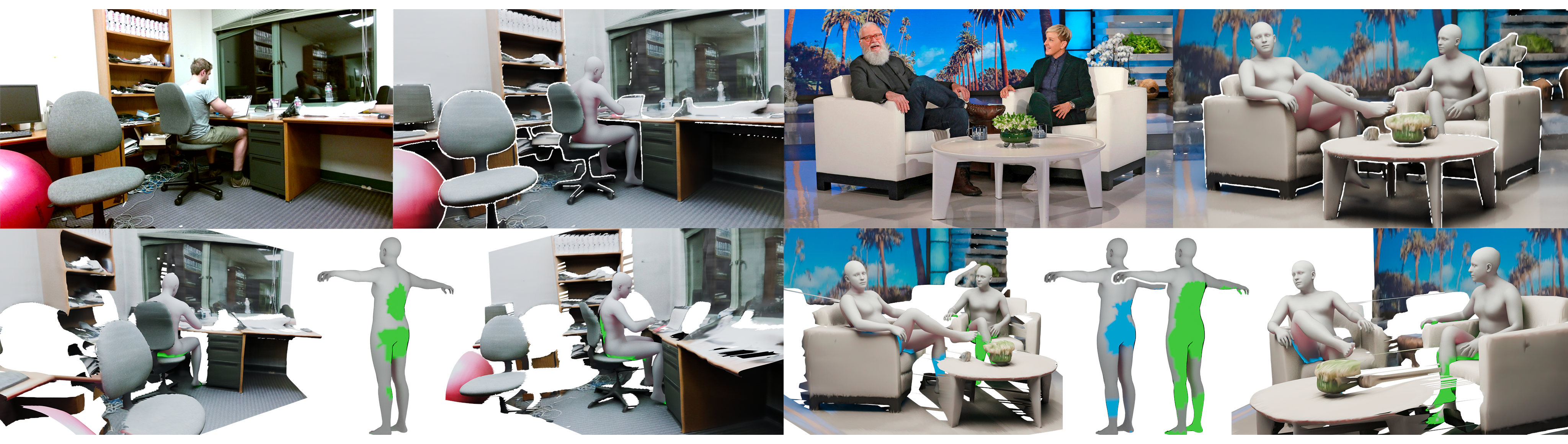}
    \caption{Given a single monocular RGB image containing a human in a complex environment, \textbf{PhySIC} reconstructs metrically aligned 3D human and scene geometries as well as a dense vertex-level contact map. Our method jointly optimizes human pose, scene geometry, and global scale to produce a physically plausible human-scene pair, accurately capturing contact and interactions such as sitting and foot-floor adherence, even in the presence of occlusions. \href{https://www.youtube.com/watch?v=HqZBMBWPhak&pp=ygUVZWxsZW4gZGF2aWQgbGV0dGVybWFu}{Image}.}
    \label{fig:teaser}
\end{flushleft}
\end{teaserfigure}

\maketitle

\input{main/01_intro}

\input{main/02_related_works}

\input{main/04_method}

\input{main/06_experiments}

\input{main/07_conclusion}

\begin{figure*}[h!]
    \centering
    \resizebox{0.9\linewidth}{!}{\includegraphics[width=0.975\linewidth]{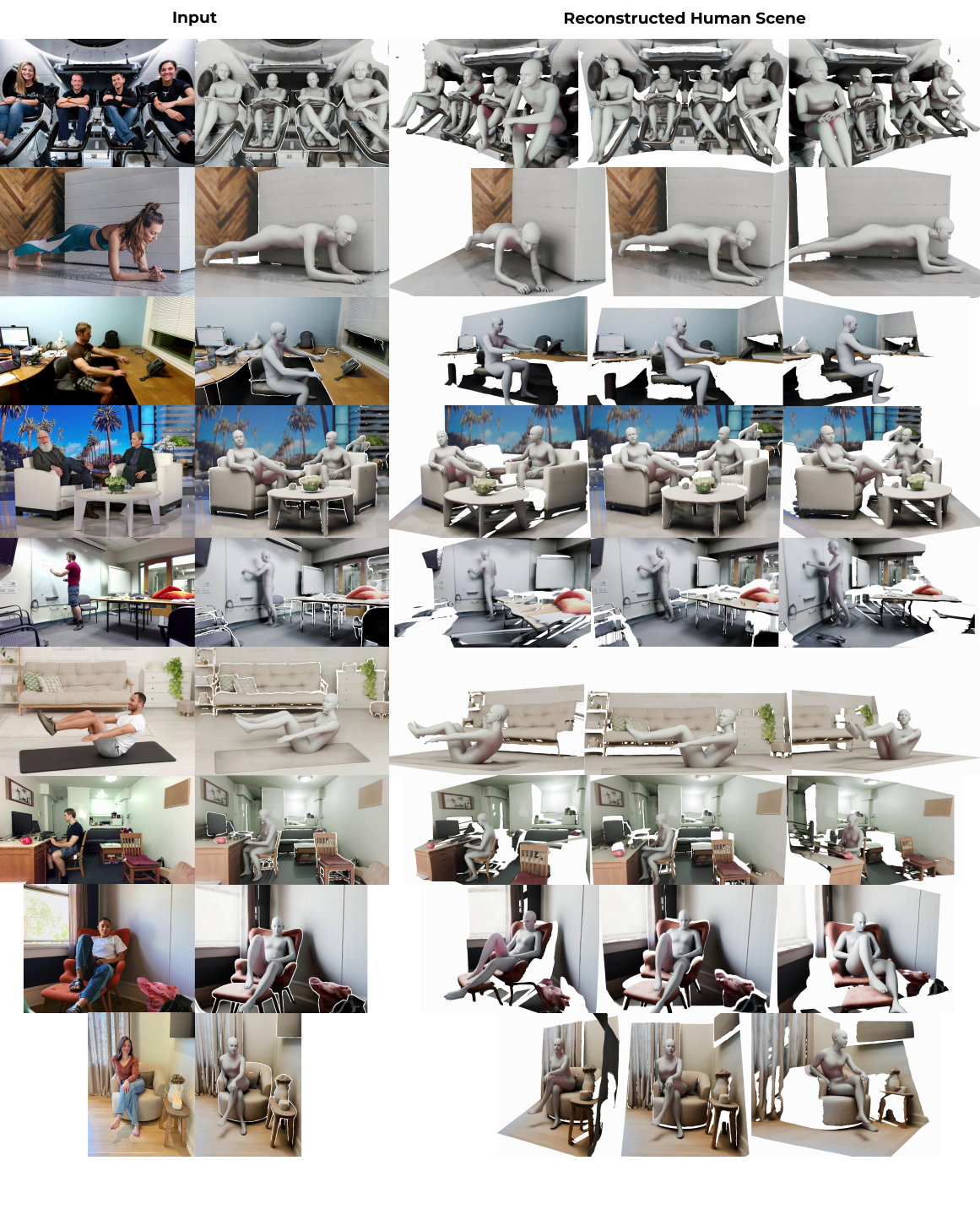}}
    \caption{\textbf{Additional qualitative results for in-the-wild images.} Please refer to Supp. Mat. for more results.} 
    \label{fig:qualitative_additional}
\end{figure*}

\begin{acks}
We thank the anonymous reviewers whose feedback helped improve this paper.
This work is made possible by funding from the Carl Zeiss Foundation. 
This work is also funded by the Deutsche Forschungsgemeinschaft (DFG, German Research Foundation) - 409792180 (EmmyNoether Programme, project: Real Virtual Humans) and the German Federal Ministry of Education and Research (BMBF): Tübingen AI Center, FKZ: 01IS18039A. 
The authors thank the International Max Planck Research School for Intelligent Systems (IMPRS-IS) for supporting YX.
PYM is supported by the Konrad Zuse School of Excellence in Learning and Intelligent Systems (ELIZA) through the DAAD programme Konrad Zuse School of Excellence in Artificial Intelligence, sponsored by the Federal Ministry of Education and Research.
GPM is a member of the Machine Learning Cluster of Excellence, EXC number 2064/1 – Project number 390727645. 

PYM and YX contributed equally as the joint first author. YX is the corresponding author. Authors with equal contributions are listed in alphabetical order and allowed to change their orders freely on their resume and website. 
YX initialized the core idea, organized the project, co-developed the current method, co-supervised the experiments, and wrote the draft. PYM co-initialized the core idea, co-developed the current method, implemented most of the prototypes, conducted experiments, and co-wrote the draft. XX contributed to the draft writing and improving Fig.~\ref{fig:method_overview}. MK lead the visualization and rendering of presented results in Figs.~\ref{fig:teaser},~\ref{fig:qualitative_additional}. %
\end{acks}

\begin{figure*}[h!]
    \centering
    \includegraphics[width=0.9875\linewidth]{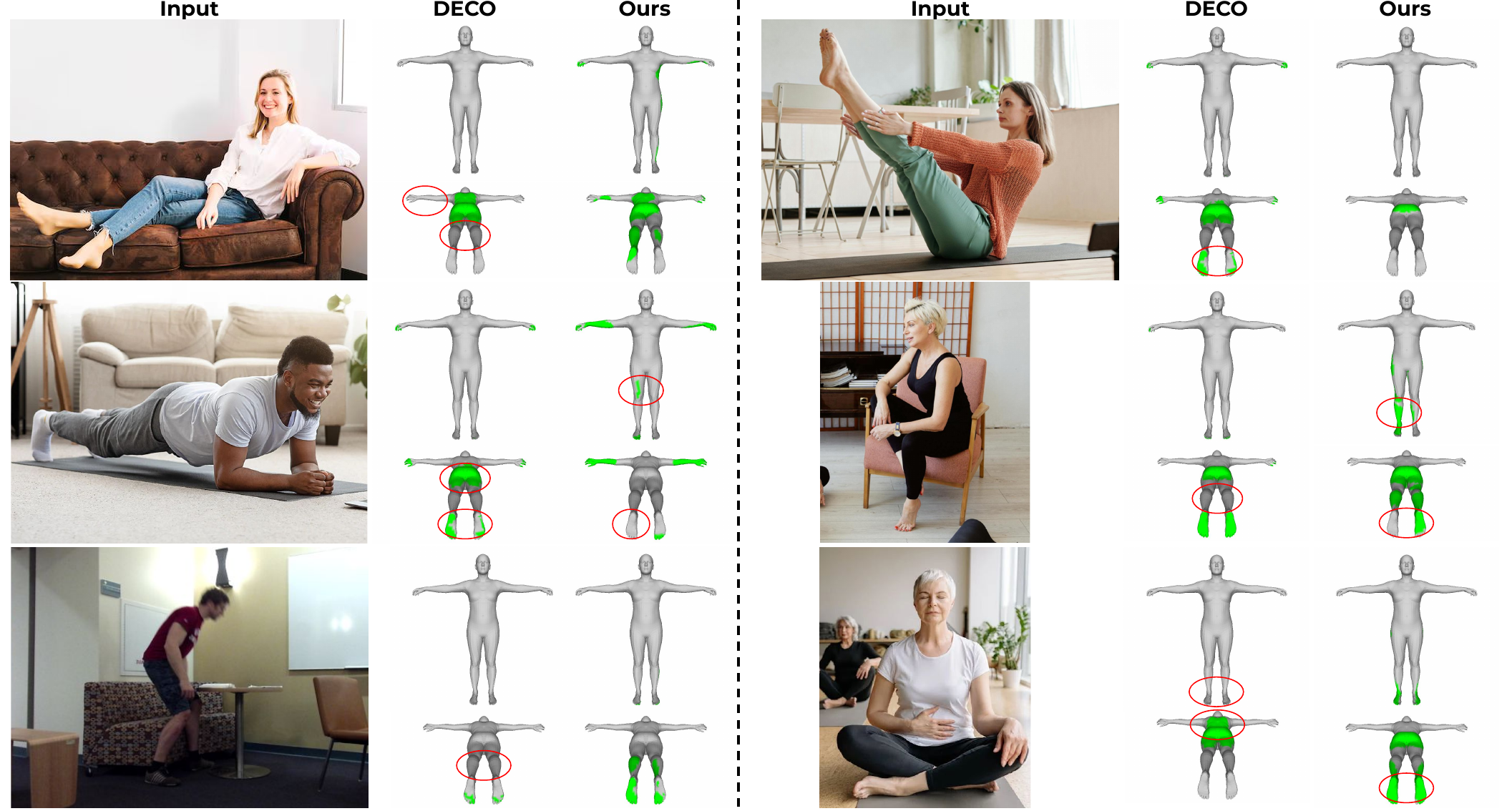} %
    \caption{\textbf{Qualitative results for contact estimation.} We compare our approach against the state-of-the-art image-based contact predictor, DECO.} 
    \label{fig:qualitative_contact_2}
\end{figure*}

\begin{figure*}[h!]
    \centering
    \resizebox{0.9\linewidth}{!}{\includegraphics[width=\linewidth]{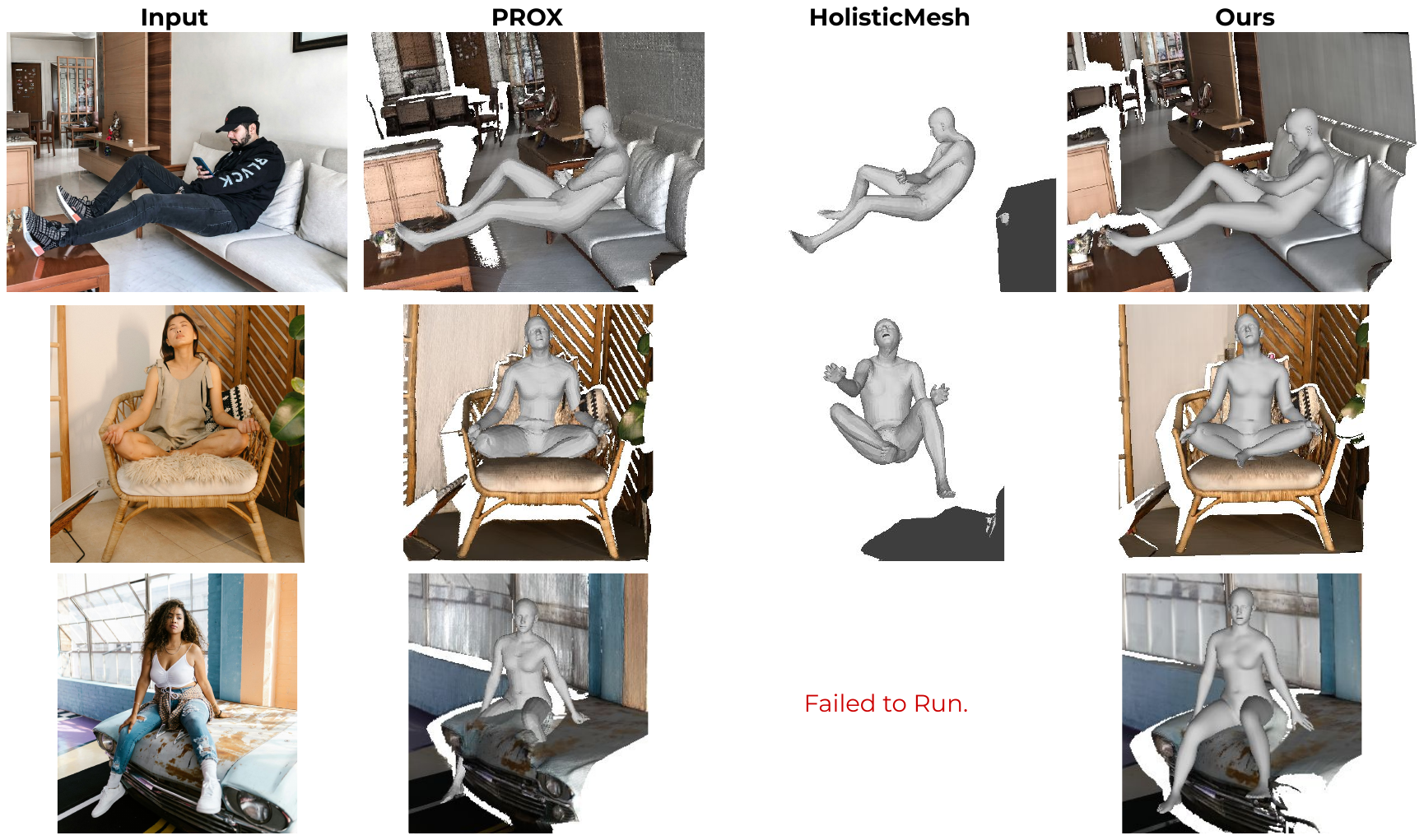}} %
    \caption{\textbf{Qualitative results on internet images.} We compare the output of \modelName{} with PROX and HolisticMesh. \newtext{HolisticMesh fails to model scenes with arbitrary surfaces since it estimates per-object geometry.}} 
    \label{fig:qualitative_recons_2}
\end{figure*}
\clearpage
\newpage 

\bibliographystyle{ACM-Reference-Format}
\bibliography{literatures}

\clearpage

\begin{appendix}
\input{main/08_supp}
\end{appendix}

\clearpage
\newpage

\end{document}

%% file: preamble.tex
\usepackage{colortbl}
\usepackage{graphicx}
\usepackage{subcaption}

\makeatletter
\DeclareRobustCommand\onedot{\futurelet\@let@token\@onedot}
\def\@onedot{\ifx\@let@token.\else.\null\fi\xspace}

\makeatother

\definecolor{yellow}{rgb}{1, 1, 0.7}
\definecolor{orange}{rgb}{1, 0.85, 0.7}
\definecolor{tablered}{rgb}{1, 0.7, 0.7}
\definecolor{red}{rgb}{1, 0, 0}

\definecolor{wincolor}{rgb}{0.85, 0.0, 0.0}

\definecolor{darkyellow}{rgb}{0.8, 0.8, 0.5}
\definecolor{darkred}{rgb}{0.7, 0.3, 0.3}
\definecolor{darkgreen}{rgb}{0.3, 0.7, 0.3}
\definecolor{green}{rgb}{0, 1.0, 0}
\definecolor{pink}{rgb}{1, 0.4, 0.7}

\definecolor{realred}{rgb}{0.95, 0.1, 0.0}

\AtEndPreamble{
    \usepackage[capitalize]{cleveref}
    \crefname{section}{Sec.}{Secs.}
    \Crefname{section}{Section}{Sections}
    \Crefname{table}{Table}{Tables}
    \crefname{table}{Tab.}{Tabs.}
}

%% file: definitions.tex
\newcommand{\pose}[0]{\boldsymbol{\theta}}
\newcommand{\shape}[0]{\boldsymbol{\beta}}
\newcommand{\trans}[0]{\boldsymbol{t}}

\newcommand{\mat}[1]{\mathbf{#1}}
\newcommand{\set}[1]{\mathcal{#1}}
\newcommand{\vect}[1]{\mathbf{#1}}

\newcommand{\modelName}{PhySIC}
\newcommand{\scene}[0]{\set{P}} %

%% file: main/01_intro.tex
\section{Introduction}
Holistic 3D understanding of humans and their surrounding environments is essential for emerging technologies such as embodied AI, sports analytics, and augmented reality. These applications require precise scene geometry, accurate localization of humans within scenes, and coherent ground contact estimation. Existing methods, however, usually consider either only static scenes without human \cite{chen2019holisticpp}, or only human pose estimation assuming given 3D scene \cite{hassan2019prox}. Recent method HolisticMesh \cite{weng_holistic_2021} can predict both scene and human from single RGB image. Nevertheless, it is limited to a single human interacting with specific indoor furniture categories, which does not scale up to arbitrary scene types. While more recent approaches like HSR~\cite{xue2024hsr} and HSfM~\cite{mueller2024hsfm} achieve holistic human-scene reconstruction, they require video input or multi-view images respectively, limiting their applicability to single-image scenarios. 

However, having a general method that can handle diverse scene types and an arbitrary number of humans interacting with a scene is very challenging. The model needs to reason about different scene geometries, intricate human-scene contacts under depth-scale ambiguity, occlusion to both human poses and scene geometry, all from a single RGB image while being fast for practical applications.

Our idea to address these challenges is to simultaneously reason about human and scene, leveraging strong geometry priors from foundation models. During interaction, the scene physically limits possible human poses and human pose provides crucial cues for estimating scene geometry and scale. Based on this observation, we propose \textbf{\modelName{}}, PhySically plausible human scene Interaction and Contacts from single RGB image. Starting from coarse monocular depth and initial parametric body estimates, our method jointly optimizes these components through an objective that harmonizes \textit{reliable depth alignment}, \textit{realistic contact encouragement}, \textit{interpenetration avoidance}, and \textit{2D reprojection consistency}, yielding coherent 3D human-scene reconstruction. In essence, PhySIC transforms a single RGB image into: (i) a metrically scaled SMPL-X human mesh, (ii) a comprehensive scene representation including dense surfaces and essential support structures like floors, and (iii) a vertex-level dense contact map within a shared metric coordinate system. Our framework is highly efficient and can process one image in \emph{less than 27 seconds}, making it possible to transform everyday images into physically consistent 3D human-scene pairs. This paves the way for scalable single-image 3D understanding. 

We evaluated \modelName{} on the PROX~\cite{hassan2019prox} and RICH~\cite{huang2022rich} datasets. Results show that our method significantly outperforms previous SOTA, HolisticMesh~\cite{weng_holistic_2021}: on PROX dataset, our method improves the mean joint error of human pose from 77mm to 42mm and contact F1 score from 0.39 to 0.51. Experiments on diverse internet images demonstrate the superior applicability of our approach to various interaction and scene types. 
Our contributions are summarized as follows:
\begin{itemize}[topsep=8pt, itemsep=6pt]
    \item We propose \modelName{}, the first metric-scale human-scene reconstruction method that can handle multiple humans, diverse scene and interaction types. 
    \item We introduce a robust initialization strategy and occlusion-aware joint optimization, providing valuable insights for human scene reconstruction. 
    \item Our highly efficient reconstruction pipeline will be publicly released, democratizing human scene reconstruction and interaction data collection. 
\end{itemize}

%% file: main/02_related_works.tex
\section{Related work}
\subsection{Single View to 3D Human}
Reconstructing the shape and pose of 3D humans from monocular images has seen significant advances, particularly with parametric models such as SMPL~\cite{loper_smpl_2015} and its extensions to SMPL-X~\cite{pavlakos_expressive_2019}, which enables expressive full-body estimation including hands and face. 
Early methods like SMPLify~\cite{bogo2016smplify} optimized body parameters to fit 2D joint detections. Subsequent deep learning methods, including HMR~\cite{kanazawa2018hmr}, SPIN~\cite{kolotouros2019spin}, and PARE~\cite{kocabas2021pare} introduced end-to-end regression and attention mechanisms to improve robustness to occlusion and truncation. WHAM~\cite{shin2024wham} and TRAM~\cite{wang2024tram} combine human mesh recovery with SLAM-based camera tracking, enabling accurate global localization of SMPL bodies in world coordinates from monocular video. 
Recently, large-scale learning-based models such as NLF~\cite{sarandi2024nlf} leverage over 25 million annotated frames to directly regress both SMPL-X parameters and global position from a single image, achieving state-of-the-art generalization and accuracy across diverse scenes and poses. Despite these advances, existing methods often lack explicit reasoning about physical interaction or consistency with the surrounding 3D scene, leading to floating, misaligned, or physically implausible human reconstructions. 
Our work addresses these issues by enabling metrically aligned, physically plausible human recovery that is explicitly consistent with the reconstructed scene.

\subsection{Single View to 3D Scene}
Early methods for monocular 3D scene reconstruction leveraged geometric and semantic priors to recover layouts, object placements, and meshes from a single RGB image. Notable among these is Total3D~\cite{nie2020total3d}, which jointly infers room layout and object pose. Mesh R-CNN~\cite{gkioxari2019meshrcnn} and MonoScene~\cite{cao2022monoscene} further advance object-centric mesh prediction and semantic scene completion. Recent breakthroughs in monocular depth estimation, such as ZoeDepth~\cite{bhat2023zoedepth}, Metric3D~\cite{hu2024metric3dv2}, and DepthPro~\cite{bochkovskiiDepthProSharp2024}, employ large-scale pretraining and transformers to predict sharp, scale-consistent depth, enabling realistic metric point cloud extraction. Gen3DSR~\cite{Ardelean2025Gen3DSR} builds on these estimators with category-specific object reconstruction, but omits human modeling and thus cannot reason about physical contact or interaction. In contrast, our method leverages state-of-the-art depth estimation together with explicit human modeling, enabling physically plausible, metrically aligned human-scene reconstruction from a single image, beyond the capabilities of prior object- or scene-centric approaches.

\subsection{3D Human-Scene Interaction}

Modeling and reconstructing plausible human–scene interactions is central to scene understanding. Early benchmarks addressed interaction detection~\cite{liu2020ntu}, generation~\cite{savva2016pigraphs}, and pose refinement~\cite{hassan_resolving_2019} with scene constraints. PROX~\cite{hassan2019prox} introduced interpenetration and contact penalties but assumes access to static scene scans. In contrast, our method reconstructs metric-scale scenes from a single RGB image. Several approaches infer scene structure from human motion~\cite{yi2022mover, nie2021pose2room, Li24PhysicPose2scene}, while large-scale capture works such as EgoBody~\cite{zhang2022egobody} and HPS~\cite{guzov2021hps} provide detailed multi-person and metric pose data using wearable sensors, though they require specialized hardware and do not address single-image reconstruction.

Dynamic tracking and contact estimation approaches, such as  CHORE~\cite{xie2022chore}, InterTrack~\cite{xie2024InterTrack}, and DECO~\cite{tripathi2023deco}, can reconstruct articulated humans and contacts, but often rely on incomplete scene geometry. Generative models like ParaHome~\cite{kim2024parahome} simulate diverse 3D human–object interactions, yet focus on activity synthesis rather than image-based reconstruction. Placement-focused works (e.g., POSA~\cite{hassan2019posa}, PLACE~\cite{zhang2020place}, Putting People in Scenes~\cite{li2019puttinghuman}) leverage statistical priors, but typically lack dense, metrically accurate scene recovery.
Recent methods in holistic reconstruction, such as RICH~\cite{huang2022rich}, HSR~\cite{xue2024hsr}, Human3R~\cite{chen2025human3r} HolisticMesh~\cite{weng_holistic_2021}, and the work by Biswas et al.~\cite{biswas2023physically} move toward integrated scene understanding but often require controlled environments. In contrast, our method reconstructs metrically accurate, physically plausible humans and diverse scenes with dense contact reasoning directly from a single image, enabling multi-human and in-the-wild scenarios (see \cref{tab:methods-comparison}).

Our work is most closely related to HSfM~\cite{mueller2024hsfm}, which reconstructs 3D human-scenes from uncalibrated multi-view images using joint optimization. To our knowledge, PhySIC is the first method to reconstruct both 3D human-scenes and their interactions from a single monocular image: a particularly challenging task due to monocular ambiguity and severe occlusions, yet highly practical given its applicability to internet images. Several additional technical design choices further differentiate PhySIC from HSfM; Please refer to our supplementary material for further details.

\begin{table}[t]
    \caption{Comparison between existing human-scene reconstruction methods and ours. Our method can handle multi-human interaction in both indoors and outdoors, and predicts the full scene at much faster speed.}
     \resizebox{0.48\textwidth}{!}{
    \begin{tabular}{l | ccccccc }
        \hline
         Method & Multi-human & Scene types & RGB input only & Output & Runtime\\
         \hline 
         PROX & \xmark & Indoor & \xmark & Objects & 73 sec. \\
         Mover & \xmark & Indoor & \xmark & Objects & 30 min. \\
         HolisticMesh & \xmark & Indoor & \cmark & Objects & 5 min. \\
         \hline
         \textit{Ours} & \cmark & In+Outdoor & \cmark & Full scene & 27 sec. \\
         \hline
    \end{tabular}}
    \label{tab:methods-comparison}
\end{table}

\begin{figure*}[t]
    \centering
    \includegraphics[width=1.0\linewidth]{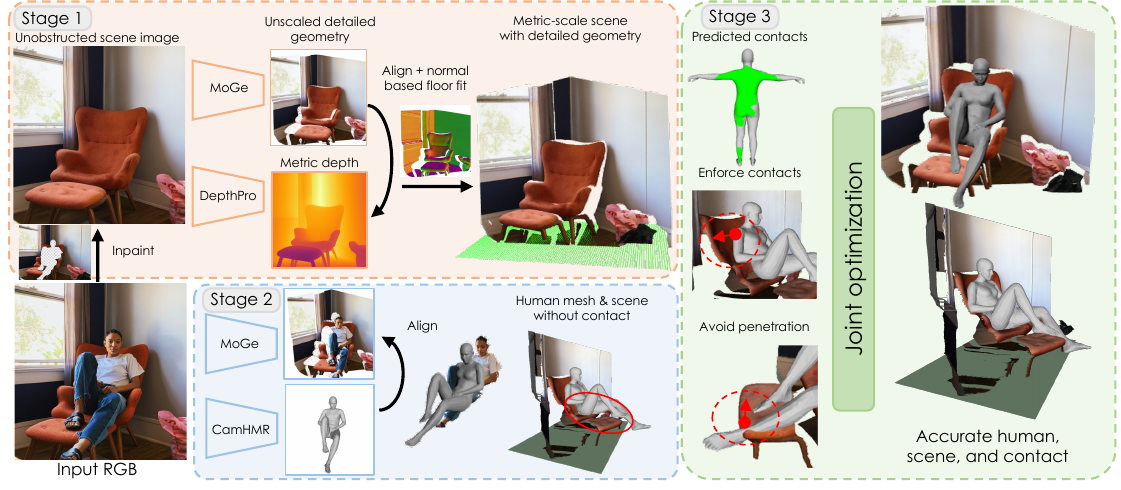}
    \caption{\textbf{Method overview.} 
    Given a single RGB image, we obtain accurate human, scene and contact reconstruction in 3D. We first obtain a complete metric scale scene with detailed geometry (Stage 1, \cref{subsec:scene-init}) and initialize human mesh which roughly aligns with the scene without contacts (Stage 2, \cref{subsec:human-init}). We then jointly optimize human and scene to satisfy contact constraints while avoiding penetrations (Stage 3, \cref{subsec:joint-opt}). \href{https://unsplash.com/photos/woman-sitting-on-sofa-chair-beside-window-Gem-93Ea_RQ}{Image from Unsplash}.}
    \label{fig:method_overview}
\end{figure*}

%% file: main/04_method.tex
\section{Method}
Given a single RGB image, our method \modelName{} predicts metric-scale dense scene point clouds and 3D human mesh with accurate vertex-level contact maps. This is a highly complex problem which requires accurate reasoning of sophisticated human poses and diverse scene geometries under heavy human-scene occlusions. We decompose this problem into separate metric-scale scene estimation (\cref{subsec:scene-init}) and human reconstruction with alignment to the scene (\cref{subsec:human-init}). Human and scene are inherently constrained by each other, which we leverage for a joint optimization to obtain physically plausible human-scene contacts (\cref{subsec:joint-opt}). An overview of our method can be found in \Cref{fig:method_overview}. 

For notational simplicity, we explain our method for single human interaction with scene, but our approach seamlessly handles multiple humans. Specifically, given input image $\set{I}\in \mathbb{R}^{H\times W\times 3}$, \modelName{} outputs scene point map $\scene_s$ and human mesh vertices $\set{V}_h\in \mathbb{R}^{N\times 3}$ using the SMPL-X body model~\cite{pavlakos_expressive_2019}. 
\subsection{Stage 1: Metric-Scale Scene with Detailed Geometry}
\label{subsec:scene-init}
\subsubsection{Scene image inpainting.} From monocular image, the human can heavily occlude the background scene, which leads to missing regions if one simply ignores the human when reconstructing the scene, causing false-negative interactions. Instead, we first inpaint the scene to fill in the missing regions and then run scene reconstruction for the complete scene, as shown in Fig.~\ref{fig:method_overview}. Specifically, we use SAM2~\cite{ravi2024sam2segmentimages} to obtain the human mask and adopt OmniEraser~\cite{weiOmniEraserRemoveObjects2025} to inpaint the human region, yielding an image $\set{I}_s$ with unobstructed view of the scene. 

\subsubsection{Metric-scale scene points.} Our goal is to obtain accurate metric-scale scene points from an image. Existing depth estimators like DepthPro~\cite{bochkovskiiDepthProSharp2024} can predict accurate metric-scale depth, however, lack detailed geometry. On the other hand, some models such as MoGe~\cite{wangMoGeUnlockingAccurate2024} can capture fine-grained details, but the results reside in relative space. We leverage the best of both worlds to obtain metric scene scale with accurate and detailed geometry. Specifically, using the inpainted scene image $\set{I}_s$, we first obtain metric depth map $\set{D}_s$ from DepthPro and unscaled relative point maps $\set{P}_s^\text{rel}$ from MoGe. Since MoGe prediction is pixel-aligned, we can align the point maps $\set{P}_s^\text{rel}$ with metric depth $\set{D}_s$ by optimizing scale $s$ and translation $\vect{t}_z$:  
\begin{equation}
    (s^*, \mathbf{t}_z^*) = \operatorname*{arg\,min}_{s, \mathbf{t}_z} \left\| (s \cdot\mathcal{\hat{P}}_s^\text{rel} + \mathbf{t}_z) - \pi^{-1}(\mathcal{D}_s, K_{\mathcal{D}}) \right\|^2_2,
    \label{eq:align_MoGe_to_dpro}
\end{equation}
where $\pi^{-1}$ is the back-projection function and intrinsic $K_{\mathcal{D}}$ is predicted by DepthPro. We optimize only the depth shift in $\vect{t}_z$ and solve this using RANSAC. The metric-scale point-map $\mathcal{\hat{P}}_s$ can then be obtained by: $\mathcal{\hat{P}}_s = s^* \cdot\mathcal{\hat{P}}_s^{rel} + \mathbf{t}_z^*.$

\subsubsection{Ground plane fitting.} The point map $\mathcal{\hat{P}}_s$ captures accurate local geometry, but can suffer from missing or unreliable floor geometry, which is important for precise human-scene interaction. To this end, we fit a plane to the floor points using normal constraints. Specifically, we adopt SAM2 to obtain a 2D mask of the floor, which is used to segment 3D floor points from $\mathcal{\hat{P}}_s$. We then use RANSAC to fit a plane robustly to the floor points, aligning both normals and positions. We estimate the normal of each point using its two immediate neighboring points defined in a 2D pixel grid. 

\subsubsection{Combined scene points.} We obtain additional floor points $\mathcal{P}_f$ by sampling a 2D grid of points on the plane within the extents of the scene. The final 3D scene, as our initialization for the next step, is formed by the union of the refined scene point cloud $\mathcal{\hat{P}}_s$ and the synthesized floor plane points $\mathcal{P}_f$:
\begin{equation}
    \mathcal{P}_{s}^\prime = \mathcal{\hat{P}}_s \cup \mathcal{P}_f.
\end{equation}

Note that the final scene points $\mathcal{P}_{s}$ come mainly from MoGe, while the initial camera used in \cref{eq:align_MoGe_to_dpro} comes from DepthPro. To ensure better alignment, we recalculate the camera intrinsic for $\mathcal{P}_{s}^\prime$. Let $(u,v)$ be a 2D pixel and $(X,Y,Z)$ be its corresponding 3D point from $\mathcal{\hat{P}}_s^\prime$, we assume centered principal point~\cite{patel_camerahmr_2024} and derive potential focal lengths: $f_x(u,v) = (u - W/2) \frac{Z}{X}$ and $f_y(u,v) = (v - H/2) \frac{Z}{Y}$. The final focal lengths, $f_x$ and $f_y$, are robustly set to the median of these respective values. This new intrinsic matrix $K$ is used for all subsequent camera projections.

\subsection{Stage 2: Human Reconstruction and Alignment}
\label{subsec:human-init}
The previous section masks out the human and only considers the scene. We now reconstruct the human and align it with the predicted scene point cloud $\scene_s^\prime$. This consists of two steps: 1) obtain human points $\set{P}_h$ aligned with scene points, and 2) estimate human mesh aligned with the human points $\set{P}_h$, i.e. the underlying scene $\scene_s^\prime$. 

\subsubsection{Metric-scale human points}. From the original input image $\set{I}$, we use MoGe to predict an unscaled point cloud $\hat{\scene}_{h+s}$, which contains human points $\hat{\scene}_{h}$ and surrounding scene points $\hat{\scene}_{s}$. We then align this to metric-scale scene points $\scene_{s}^\prime$ by optimizing a scale and depth shift, similar to \cref{eq:align_MoGe_to_dpro}. Note that we use the human mask to remove $\hat{\scene}_{h}$ from $\hat{\scene}_{h+s}$ when performing the alignment. We then apply the optimized scale and shift to human points $\hat{\scene}_{h}$, thus aligning them with the metric-scale scene, denoted as $\scene_{h}$.

\subsubsection{Human mesh estimation}. To obtain semantically meaningful contact vertices, we use SMPL-X~\cite{pavlakos_expressive_2019} to represent the human. We denote $\mat{H}$ as the SMPL-X model which takes body shape $\shape$, hand and full body poses $\pose_h, \pose_b$, and global translation $\trans_h$ as input, and outputs the human vertices $\set{V}_h=\mat{H}(\shape, \pose_h, \pose_b, \trans_h)$. The initial SMPL-X vertices $\set{V}_h$ are obtained by fusing SMPL~\cite{loper_smpl_2015} prediction from CameraHMR~\cite{patel_camerahmr_2024} and hand pose from WiLor~\cite{potamias_wilor_2025}. Specifically, we fit SMPL-X into the SMPL mesh predicted by CameraHMR using SMPLFitter~\cite{sarandi2024nlf} and replace the hand parameters with the hand pose predicted by WiLor. This initial estimation does not precisely align with the input image and metric-scale scene, which we address next. 

\subsubsection{Metric-scale human mesh}. We first optimize the global human translation $\trans_h$ to improve the pixel-alignment of the estimated SMPL-X vertices using 2D joint projection loss: 
\begin{equation}
    L_{\text{j2d}} = \left\| \left( \pi(\mathcal{J}(\set{V}_h(\mathbf{t}_h), K) - \hat{J}_h^{2D} \right) \right\|_2^2,
    \label{eq:joint-projection-loss}
\end{equation}
where $\set{J}: \mathbb{R}^{N\times 3}\mapsto \mathbb{R}^{J\times 3}$ regresses the 3D body keypoints and $\hat{J}_h^{2D}$ are the 2D keypoints predicted by ViTPose~\cite{xu2022vitposesimplevisiontransformer}. We then align the optimized human vertices with the metric-scale human points $\scene_{h}$ using the Chamfer distance between camera-facing vertices $\mathcal{V}_{\text{cf}}$ and human points $\scene_{h}$: 
\begin{align}
    L_\text{align} &= \lambda_\text{j2d} L_\text{j2d} + \lambda_\text{d} L_\text{d}, \text{  where} \\
    L_\text{d} &=\sum_{\vect{v} \in \mathcal{V}_{\text{cf}}} \min_{\vect{p} \in \mathcal{P}_h} \|\vect{v} - \vect{p}\|_2^2 + \sum_{\vect{p} \in \mathcal{P}_h} \min_{\vect{v} \in \mathcal{V}_{\text{cf}}} \|\vect{p} - \vect{v}\|_2^2.
    \label{eq:align-vertices-points}
\end{align}
We select camera-facing vertices $\mathcal{V}_{cf} \subset \mathcal{V}_h$ as the vertices whose surface normals are at an angle deviating less than 70 degrees from the camera view direction. This is crucial to avoid aligning the backside vertices with the human points. Note here that we only optimize the global translation $\trans_h$ parameter.

\subsection{Stage 3: Joint Human-Scene Optimization}
\label{subsec:joint-opt} 
Even though the human vertices $\set{V}_h$ and metric-scale scene points $\scene_s^\prime$, which were obtained from previous steps, reside in the same metric-scale coordinate, they are predicted separately. Hence, physical plausibility is not guaranteed. We further enhance the plausibility by enforcing additional constraints between the human and the scene (Fig.~\ref{fig:method_overview}, Stage 3). To this end, we formulate a joint optimization objective that adapts principles of contact attraction and interpenetration avoidance~\cite{hassan2019prox, yi2022mover} to our setting of single-image reconstruction with pointmaps. Thus, we additionally introduce the contact and interpenetration loss, together with regularization terms to jointly optimize the human parameters $\pose_b, \pose_h, \shape, \trans_h$ and a scene scale parameter $s_\text{sc}$:
\begin{equation}
    L_\text{total} = \lambda_\text{j2d} L_\text{j2d} + \lambda_\text{d} L_\text{d} + \lambda_c L_c + \lambda_i L_i + \lambda_\text{reg} L_\text{reg}.
    \label{eq:loss_total}
\end{equation}
Let $\scene_s=s_\text{sc} \scene_s^\prime$ be the scaled scene points; we explain the contact, interpenetration and regularization terms next. The loss weights $\lambda_*$ are detailed in the supplementary. 

\subsubsection{Contact loss $L_c$.} This encourages the human vertices in contact with the scene to be close to the scene points $\scene_s$. We use DECO~\cite{tripathi2023deco}, which predicts the human contact vertices $\set{V}_\text{con}$ and minimize their distance to the closest scene points. During optimization, we use an active-contact subset by re-evaluating nearest scene distances each iteration and only applying $L_c$ to vertices within $\epsilon$, suppressing spurious long-range contacts:

\begin{equation}
    L_{\text{c}} = \sum_{v \in \mathcal{V}_{\text{con}}} \left( \min_{\vect{p} \in \mathcal{P}_s} \rho (\|\vect{v} - \vect{p}\|_2^2) \right) \mathbb{I} \left( \min_{\vect{p} \in \mathcal{P}_s} \|\vect{v} - \vect{p}\|_2^2 < \epsilon \right),
    \label{eq:loss_contact}
\end{equation}
where $\rho$ is an adaptive robust loss function~\cite{BarronCVPR2019} and the indicator function $\mathbb{I}(\cdot)$ ensures the loss term is only active when the distance to the nearest scene point is less than a threshold $\epsilon$. This hinders penalizing distant false-positive contact predictions or interactions with outlier scene points.

\subsubsection{Occlusion aware interpenetration loss $L_i$.} It prevents the human mesh $\mathcal{V}_h$ from unnaturally penetrating the scene geometry $\mathcal{P}_s$. We leverage the estimated per-point normal of $\mathcal{P}_s$ and penalize points lying opposite to the normal direction:
\begin{equation}
    L_{\text{i}} = \sum_{\vect{v} \in \mathcal{V}_h \setminus \mathcal{V}_{\text{occ}}} \rho \left( \min_{\vect{p} \in \mathcal{P}_s} \|\vect{v} - \vect{p}\|_2^2 \right) \mathbb{I}(\vect{n}_\vect{p} \cdot (\vect{v} - \vect{p}) < 0).
    \label{eq:loss_penetration}
\end{equation}
Importantly, we exclude human vertices $\mathcal{V}_{\text{occ}}$ occluded by surrounding objects or by itself. Specifically, we consider human vertices whose 2D projections lie outside the human mask as occluded by object. We divide the vertices into different body parts and consider a part as self-occluded if 30\% of its vertices are occluded by other body parts. This prevents the occluded body parts from moving towards unnatural poses due to the penetration loss, as no other signal, like 2D keypoints, is available to regularize the optimization. %

\subsubsection{Regularization terms $L_\text{reg}$.} To ensure the optimized human mesh $\mathcal{V}_h$ does not deviate excessively from initial estimates, we apply a mesh regularization loss, treating the initial estimates as a pose prior. This loss penalizes the L2 distance between the current and initial mesh vertices in the root-relative space, constraining the local body pose of the human, while allowing for large updates in global translation of the mesh. We increase the weight of the regularization loss for occluded vertices $\mathcal{V}_{occ}$ since the initial estimates are our best guess for unobserved parts of the human mesh. We further weakly regularize the scene scale $s_{sc}$ and the human translation $\trans_h$ by preventing large deviations from their initial values.

\subsubsection{Contact map extraction.} Our joint optimization produces accurate and physically plausible human-scene interactions, which allows us to extract per-vertex contact maps based on proximity. Each human mesh vertex $\mathbf{v}_j \in \mathcal{V}_h$ is labeled as in-contact if its Euclidean distance to the nearest point on the scene surface is less than a predefined threshold $\epsilon_c$. This process yields a binary contact mask over the human mesh vertices, identifying regions of interaction.

\subsubsection{Handling multiple humans.} The method described above can easily be extended to multiple humans by using another human mask to perform human-scene alignment and joint optimization. Specifically, we use SAM2 to obtain per-instance human masks. We inpaint all humans simultaneously to obtain the scene, then align each human mesh individually with the scene following \cref{subsec:human-init}. We then perform one joint optimization between the underlying scene and all humans using \cref{eq:loss_total}. 

\begin{figure*}[t]
    \centering
    \resizebox{0.9\linewidth}{!}{\includegraphics[width=0.95\linewidth]{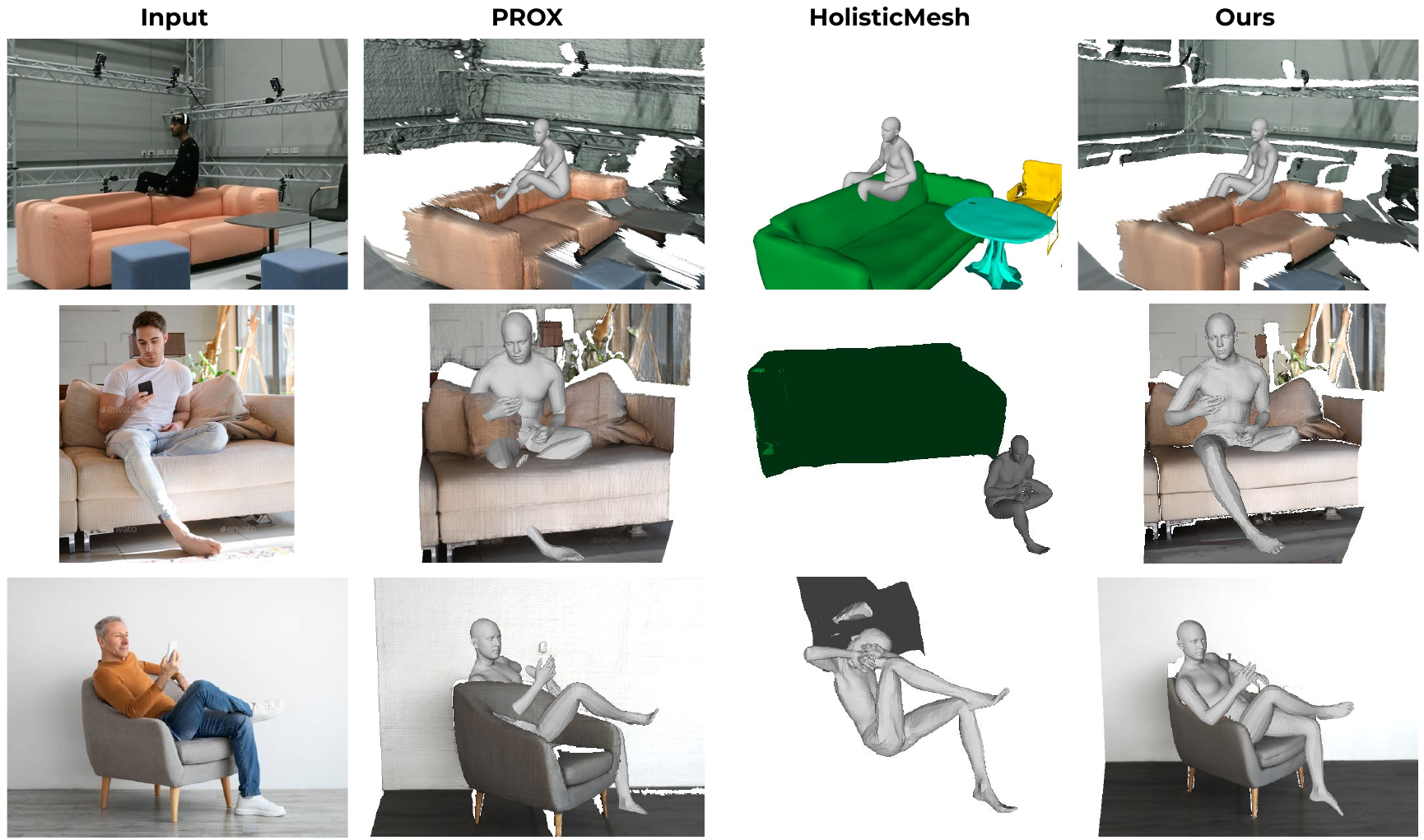}} %
    \caption{\textbf{Qualitative results on PROX dataset (row 1) and internet images (row 2-3).} We compare the output of \modelName{} with PROX \cite{hassan2019prox} and HolisticMesh \cite{weng_holistic_2021}. Note that we run PROX with our estimated scene on internet images as there is no scene scan available. Both PROX and HolisticMesh rely on predefined contact maps, hence are not robust to complex human poses and interactions. Our method reconstructs 3D scene and adapts contact optimization based on input, leading to more coherent reconstruction. Please refer Fig.~\ref{fig:qualitative_additional} and Fig.~\ref{fig:qualitative_recons_2} for more results.
    } 
    \label{fig:qualitative_recons}
\end{figure*}

%% file: main/06_experiments.tex
\section{Experiments}

\subsection{Implementation Details}
We implement our optimization framework using PyTorch~\cite{paszke2019pytorch} and batched 3D geometry operations to handle multiple humans with PyTorch3D~\cite{ravi2020pytorch3d}. During initialization, we perform aggressive outlier point removal using mean $k$-NN distance to ensure clean scene geometry, where $k$ is adaptively set based on image resolution. For the first optimization (Eq.~\ref{eq:joint-projection-loss}), we perform 30 iterations of gradient descent with Adam~\cite{kingma2017adam}. For the second optimization (Eq.~\ref{eq:align-vertices-points}), we use two iterations of L-BFGS~\cite{Liu1989OnTL}. Our final optimization (Eq.~\ref{eq:loss_total}) utilizes 100 iterations of gradient descent with Adam. Both gradient descents use a learning rate of $1e$-$2$, and the L-BFGS optimizer uses a unit learning rate. While the camera-facing mask $\mathcal{V}_{cf}$ remains stable throughout optimization, the self-occlusion state can vary due to pose optimization. Hence, we update $\mathcal{V}_{occ}$ every 30 iterations of the final gradient descent. \newtext{For more details, please refer supplementary.}

A frequent operation used in $L_c$ and $L_i$ is nearest-neighbor search. Despite the varying scene scale during optimization, we leverage the scale-invariant nature of the nearest neighborhood structure to pre-compute a $128^3$ grid of nearest-scene-points, and transform query points to the initial scale. This results in a $15\text{–}20\times$ overall speedup, compared to a brute-force implementation. On a NVIDIA H100 GPU, our optimization takes 9 seconds for a 480p image and 12 seconds for a 720p image, yielding end-to-end human-scene reconstruction times of 27 seconds and 36 seconds respectively.

\begin{table}[h] %
\centering
\caption{\textbf{Quantitative Comparison on PROX and RICH.} Our method outputs better local pose (PA-MPJPE) and more accurate contacts. Although HolisticMesh shows better number in MPJPE on PROX, we found it is unreliable and cannot robustly reconstruct on in-the-wild data (See Fig.~\ref{fig:qualitative_recons}).}
\label{tab:quantitative_comparison}
\resizebox{0.96\columnwidth}{!}{
    \begin{tabular}{l S[table-format=2.2] S[table-format=3.2] S[table-format=3.2] *{3}{S[table-format=1.3]}}
    \toprule
    \multirow{2}{*}{\textbf{Method}} & \multicolumn{3}{c}{\textbf{Human Pose Metrics} $\downarrow$} & \multicolumn{3}{c}{\textbf{Contact Metrics} $\uparrow$} \\
    \cmidrule(lr){2-4} \cmidrule(lr){5-7} %
    & {PA-MPJPE} & {MPJPE} & {MPVPE} & {Precision} & {Recall} & {F1 score} \\
    \midrule
    \multicolumn{7}{l}{\textit{PROX Quantitative Dataset}} \\ %
    \addlinespace[0.5em]
    CameraHMR & \cellcolor{heatOrange}42.35 & 997.49 & 996.20 & {--} & {--} & {--} \\
    DECO & {--} & {--} & {--} & \cellcolor{heatOrange}0.406 & \cellcolor{heatYellow}0.349 & \cellcolor{heatYellow}0.376 \\
    PROX         & \cellcolor{heatYellow}73.31   & \cellcolor{heatYellow}266.50  & \cellcolor{heatYellow}266.00  & 0.260   & 0.108  & 0.152 \\
    HolisticMesh & 77.04   & \cellcolor{heatRed}{\textbf{202.80}} & \cellcolor{heatRed}{\textbf{191.70}} & \cellcolor{heatYellow}0.373   & \cellcolor{heatOrange}0.412  & \cellcolor{heatOrange}0.391 \\
    \textit{Ours} & \cellcolor{heatRed}{\textbf{41.99}} & \cellcolor{heatOrange}230.26  & \cellcolor{heatOrange}227.19  & \cellcolor{heatRed}{\textbf{0.508}} & \cellcolor{heatRed}{\textbf{0.514}} & \cellcolor{heatRed}{\textbf{0.511}} \\
    \addlinespace[0.55em]
    \hline
    \multicolumn{7}{l}{\textit{RICH-100 Dataset}} \\ %
    \addlinespace[0.5em]
    PROX         & \cellcolor{heatOrange}120.24   & \cellcolor{heatOrange}706.19  & \cellcolor{heatOrange}692.07  & \cellcolor{heatOrange}0.040   & \cellcolor{heatOrange}0.250  & \cellcolor{heatOrange}0.069 \\ %
    \textit{Ours} & \cellcolor{heatRed}{\textbf{46.50}} & \cellcolor{heatRed}{\textbf{616.27}} & \cellcolor{heatRed}{\textbf{617.33}} & \cellcolor{heatRed}{\textbf{0.310}} & \cellcolor{heatRed}\textbf{0.689} & \cellcolor{heatRed}\textbf{0.428} \\ %
    \bottomrule
    \end{tabular}
}
\end{table}

\subsection{Evaluation Protocol}

We evaluate \modelName{} against prior arts on the PROX~\cite{hassan_resolving_2019} and RICH~\cite{huang2022rich} datasets, both containing humans and static scene scans. \newtext{The PROX dataset captures a single subject interacting with various objects of a scene in an indoor setting. In contrast, RICH includes videos of two scenes covering both indoor and outdoor settings, each scene captured by $6$-$8$ cameras, resulting in $\sim125k$ frames, with high redundancy, which makes full evaluation expensive. Hence, } we use all 178 images from PROX-quantitative and randomly sample 100 images from RICH, covering all possible \newtext{cameras, activities and} backgrounds. We further provide qualitative results on the PiGraphs dataset~\cite{savva2016pigraphs}, containing videos of humans interacting with static scenes in indoor environments. Finally, we collect a set of in-the-wild images from the internet to show the generalizability of our approach.

We compare \modelName{} with PROX~\cite{hassan_resolving_2019} and HolisticMesh~\cite{weng_holistic_2021}, two state-of-the-art approaches that jointly model human-scene interaction from monocular images. While HolisticMesh estimates human-scenes from a single RGB image, whereas PROX requires a static 3D scene scan for optimization. To enable a fair comparison, and to evaluate PROX on RGB images, we perform two modifications. First, we replace the static scene with an unprojected depth map from DepthPro~\cite{bochkovskiiDepthProSharp2024}. Further, we replace PROX's pose prior VPoser~\cite{pavlakos_expressive_2019} with SOTA CameraHMR. Specifically, we initialize and regularize the pose optimization using CameraHMR~\cite{patel_camerahmr_2024}. 

Unlike HolisticMesh, which fits a single static scene to the entire PROX-Quantitative sequence, our method relies only on a single inpainted image from a frame without interactions with thin structures. This avoids the need for per-frame inpainting and allows us to optimize the human and the scene independently at each frame, without relying on sequence-level cues. Despite this lightweight design, our inpainting generalizes robustly and works well on in-the-wild images directly, even without any sequence information.

\begin{figure*}[t]
    \centering
    \includegraphics[width=0.9\linewidth]{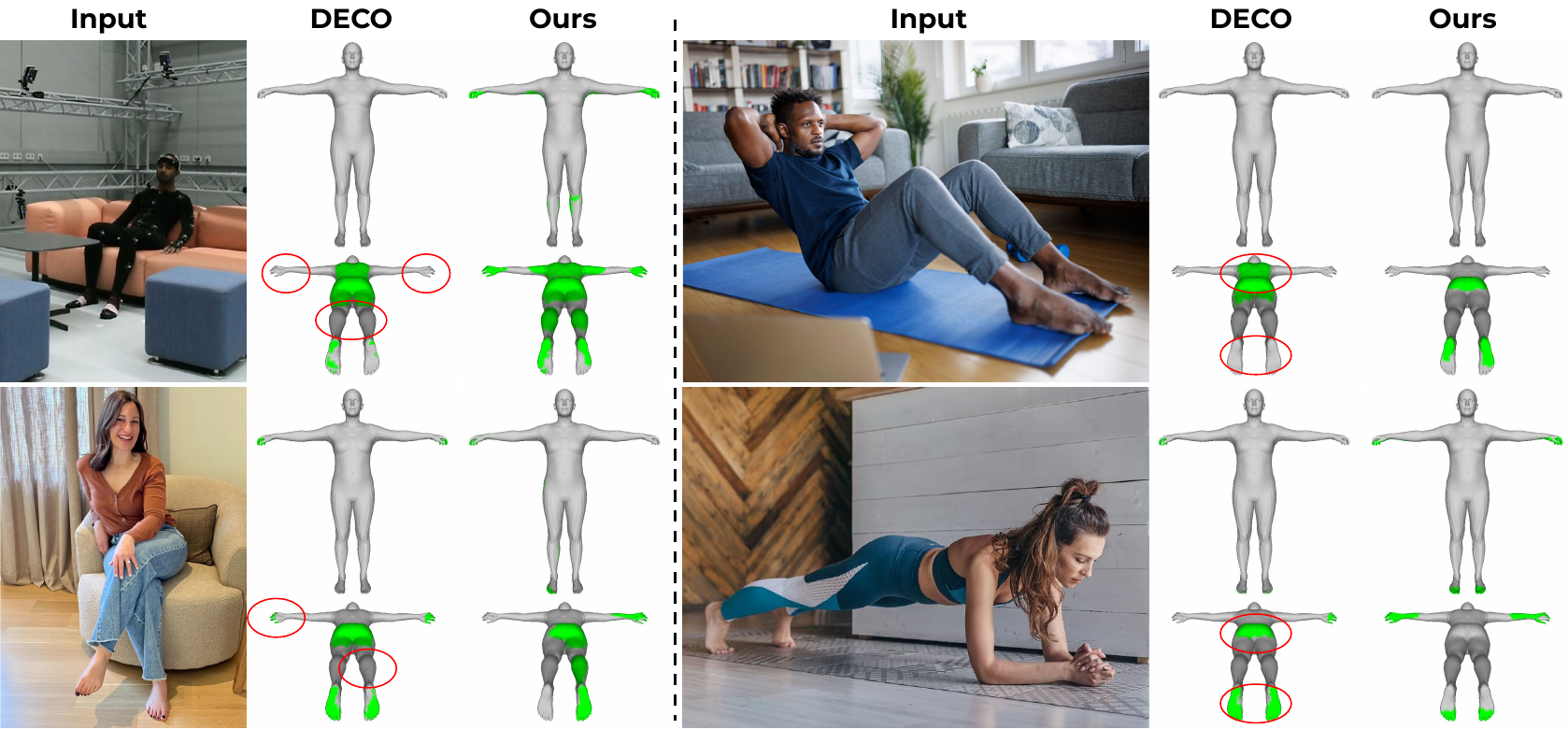} %
    \caption{\textbf{Qualitative results for contact estimation.} We compare our approach against the state-of-the-art image-based contact predictor, DECO \cite{tripathi2023deco} in both lab and wild setting. Note how our method improves the nuanced contact on arms and feet. Please refer Fig.~\ref{fig:qualitative_contact_2} for further examples.} 
    \label{fig:qualitative_contact}
\end{figure*}

\subsection{Qualitative Analysis}
\subsubsection{Human-scene reconstruction.} ~\cref{fig:qualitative_recons} presents qualitative results of human-scene reconstruction. In contrast to our method, PROX lacks robust occlusion handling and an appropriate distance threshold in its contact loss, resulting in inaccurate poses and mislocalization within the scene. While some interpenetration is expected due to unmodeled scene deformations, PROX exhibits excessive interpenetration beyond these expected discrepancies. Similarly, HolisticMesh also suffers from noticeable interpenetration, and fails to run on certain in-the-wild examples, highlighting limitations with generalizability. In contrast, \modelName{}'s robust occlusion handling and refined distance thresholding for contacts leads to more accurate poses, better localization, and significantly reduced interpenetration, thereby successfully increasing robustness in complex scenes. For additional results, please refer to Fig.~\ref{fig:qualitative_recons_2} \newtext{and supplementary}.

\subsubsection{Contact estimation.} We show example comparison with contact estimation method DECO~\cite{tripathi2023deco} in \cref{fig:qualitative_contact}. Our joint optimization is guided by the contact estimation from DECO which can be noisy. However, our approach robustly recovers accurate human-scene interactions and further improves contacts, especially in the intricate body parts such as feet and arms. Additional examples can be found in Fig.~\ref{fig:qualitative_contact_2}.

\subsection{Quantitative Analysis}

We quantitatively evaluate our method on both 3D human pose and vertex-level contact metrics. For 3D human pose, we report the Mean Per-Joint Position Error (MPJPE), the average euclidean distance between the camera-relative predicted and GT human joints. We also use Procrustes Aligned MPJPE (PA-MPJPE) which computes MPJPE after global alignment, effectively comparing the root-relative human pose. Additionaly, we report the Mean Per-Vertex Positon Error (MPVPE), the average Euclidean distance between the predicted and GT mesh vertices, which takes into account the predicted human shape $\shape$. For the human-scene contact, we report standard classification metrics (precision, recall, F1 score), calculated using the predicted and GT per-vertex contacts.

The results in Tab.~\ref{tab:quantitative_comparison} demonstrate that our approach achieves state-of-the-art performance in both human pose and contact estimation. Specifically on the PROX dataset, we significantly reduce PA-MPJPE, by nearly half compared to the PROX and HolisticMesh, even though both methods initialize from CameraHMR. Our method consistently improves upon the state-of-the-art CameraHMR and DECO across all pose and contact metrics. Our method also outperforms HolisticMesh in contact accuracy with a 40$\%$ improvement in F1 score. Although HolisticMesh shows a marginally better MPJPE and MPVPE on PROX, it performs poorly on non-PROX in-the-wild images (Fig.~\ref{fig:qualitative_recons_2}), and suffers from severe interpenetration and inaccurate local poses, as indicated by its PA-MPJPE.

On the RICH dataset, HolisticMesh could not be evaluated, as it is trained only for indoor living environments occupied with limited object categories, while RICH was captured both indoors and outdoors, beyond HolisticMesh categories. Our method outperforms PROX on both pose and contact metrics.

\subsection{Ablation Study}
We investigate the impact of different loss terms on our joint optimization stage and report the results on PROX dataset. Starting from the basic 2D joint reprojection loss $L_\text{j2d}$ and regularization term, we gradually add more losses defined in \cref{eq:loss_total} to the optimization process. We report the performance of our initialization approaches, CameraHMR and DECO, and ablation results in \cref{tab:ablation_results}. \newtext{With just $L_{reg} + L_{j2d}$, the human pose metrics degrade compared to the initial estimates. This is due to depth ambiguity in the monocular setting: a perfect 2D fit does not imply accurate 3D pose. This necessitates additional constraints from human-scene interaction losses ($L_c$ and $L_i$). However, since these losses are applied against nearest scene points, they can misalign with the actual contact regions. To address this, we also include a loss against the human points $\set{P}_h$.}
The depth alignment loss ($+L_{d}$) marks a crucial improvement in both pose and contacts, due to improved localization of the human within the scene\newtext{, which ensures that $L_c$ and $L_i$ act on the correct scene regions and thus become effective}. Our occlusion-aware interpenetration loss ($L_i$ with $\set{V}_\text{occ}$ excluded) further then delivers the largest PA-MPJPE gain (to 41.91) and achieves the best contact recall and F1 scores. Note that without the occlusion awareness, the local body poses (PA-MPJPE) are even worse than the initialization. This is because the occluded part can be overly penalized due to interpenetration without regularization from the input image, leading to large deviations from correct body poses. Our experiments highlight the significance of occlusion reasoning, which is consistent with the prior works~\cite{xie2023vistracker, xie2024InterTrack}. For detailed qualitative results, please refer to supplementary.

\begin{table}[t]
\centering
\caption{Ablating the impact of different loss terms on joint human-scene optimization. Depth loss $L_d$ is important to ensure good global alignment (MPJPE) and occlusion-aware interpenetration loss $L_i$ improves local pose accuracy (PA-MPJPE).}
\label{tab:ablation_results}
\resizebox{0.96\columnwidth}{!}{
    \begin{tabular}{l S[table-format=2.2] S[table-format=3.2] S[table-format=3.2] *{3}{S[table-format=1.3]}}
    \toprule
    \multirow{2}{*}{\textbf{Ablation}} & \multicolumn{3}{c}{\textbf{Human Pose Metrics} $\downarrow$} & \multicolumn{3}{c}{\textbf{Contact Metrics} $\uparrow$} \\
    \cmidrule(lr){2-4} \cmidrule(lr){5-7} %
    & {PA-MPJPE} & {MPJPE} & {MPVPE} & {Precision} & {Recall} & {F1 score} \\
    \midrule
    Init. (CHMR)             & \cellcolor{heatYellow}42.35   & 997.49  & 996.20  & {--}     & {--}    & {--}    \\ %
    Init. (DECO)             & {--}    & {--}    & {--}    & \cellcolor{heatYellow}0.406   & 0.349  & 0.376 \\ %
    \hline
    \addlinespace[0.5em] %
    $L_{reg} + L_{j2d}$ & 76.79   & 643.02  & 641.31  & 0.288   & 0.052  & 0.088 \\ %
    $+ L_{c}$  & 71.22   & 637.50  & 635.98  & 0.397   & 0.250  & 0.307 \\ %
    $+ L_{i}$   & 69.22   & 639.16  & 637.62  & 0.394   & 0.228  & 0.289 \\ %
    $+ L_{d}$    & 72.64   & \cellcolor{heatYellow}364.98  & \cellcolor{heatYellow}358.75  & \cellcolor{heatOrange}0.490   & \cellcolor{heatYellow}0.430  & \cellcolor{heatYellow}0.459 \\ %
    $+$ occ. aware $L_i$  & \cellcolor{heatRed}{\textbf{41.91}} & \cellcolor{heatOrange}238.72  & \cellcolor{heatOrange}235.75  & \cellcolor{heatOrange}0.490   & \cellcolor{heatRed}{\textbf{0.550}} & \cellcolor{heatRed}{\textbf{0.518}} \\ %
    $+$ floor (full model)        & \cellcolor{heatOrange}41.99   & \cellcolor{heatRed}{\textbf{230.26}} & \cellcolor{heatRed}{\textbf{227.19}} & \cellcolor{heatRed}{\textbf{0.508}} & \cellcolor{heatOrange}0.514  & \cellcolor{heatOrange}0.511 \\ %
    \bottomrule
    \end{tabular}
}
\end{table}

%% file: main/07_conclusion.tex
\section{Limitations and Future Works}
While PhySIC advances the state of the art in physically plausible human-scene reconstruction from a single image, several limitations remain, highlighting directions for future research. (i) {Image inpainting.} Our approach relies on state-of-the-art inpainting models~\cite{weiOmniEraserRemoveObjects2025} to reconstruct occluded scene regions. These models are imperfect, particularly for thin or intricate structures, leading to erased or deformed geometry. As inpainting methods improve, we expect direct benefits to our approach. (ii) {Scene deformations.} We assume static, rigid scene geometry, which may not hold for deformable objects such as cushions or clothing. Extending \modelName{} to handle non-rigid scene deformations could enable more realistic human-scene interactions. (iii) {Human-object interactions.} We focus on human-scene interactions and do not explicitly model fine-grained interactions with small objects, such as grasping or pushing. Future work could integrate off-the-shelf object mesh estimators, align them with reconstructed depth maps, and leverage additional 2D object supervision. (iv) {Flat floor assumption.} \modelName{} assumes that floors are planar to simplify occlusion reasoning and contact estimation. This assumption generally holds, but can fail when no floor points are detected or RANSAC fails to find consensus. In such cases, we skip floor sampling, which may result in false-negative contacts. Maturity of holistic 3D scene reconstruction methods could address this limitation~\cite{ roh_catsplat_2024}.

\section{Conclusion}

We present \modelName{}, a framework for physically plausible human-scene interaction and contact reconstruction from a monocular RGB image. By jointly optimizing metrically scaled SMPL-X human meshes and detailed 3D scene geometry, \modelName{} enables reconstruction of coherent, physically realistic human-scene pairs in diverse environments. Our method introduces robust initialization strategies, combining metric depth and detailed relative geometry, occlusion-aware refinement, and efficient multi-term optimization that enforces contact, interpenetration, and depth alignment. Extensive experiments on challenging benchmarks demonstrate that \modelName{} significantly outperforms prior work in both pose and contact metrics, and generalizes well to multi-human and in-the-wild scenarios.
\modelName{} provides a scalable, accessible step towards holistic, single-image 3D human-centric scene understanding. We anticipate that continued progress in inpainting, foundation geometry models, and interaction reasoning will further enhance the capabilities and generality of our approach. We will release our code and evaluation scripts to support future research and practical applications.

%% file: main/08_supp.tex
\renewcommand{\thefigure}{S\arabic{figure}}

\begin{figure*}[h!]
    \centering
    \includegraphics[width=\linewidth]{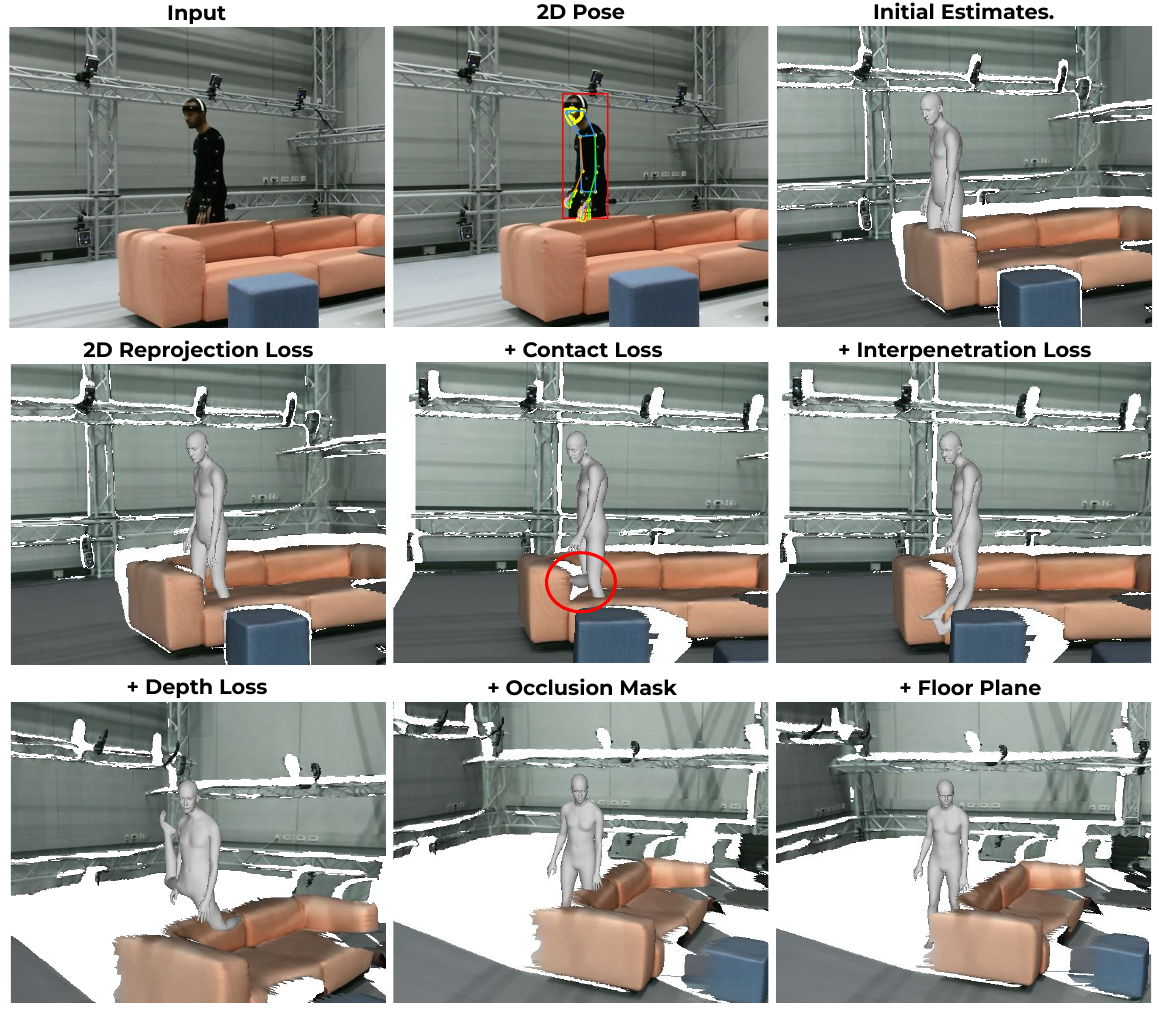} %
    \caption{\textbf{Reconstructions under ablation.} We perform a qualitative ablation study to evaluate various design choices; details and full results are provided in  Sec. F.}
    \label{fig:occlusion_ablation}
\end{figure*}

\section{Additional Results }
\begin{figure*}[h!]
    \centering
    \includegraphics[width=0.95\linewidth]{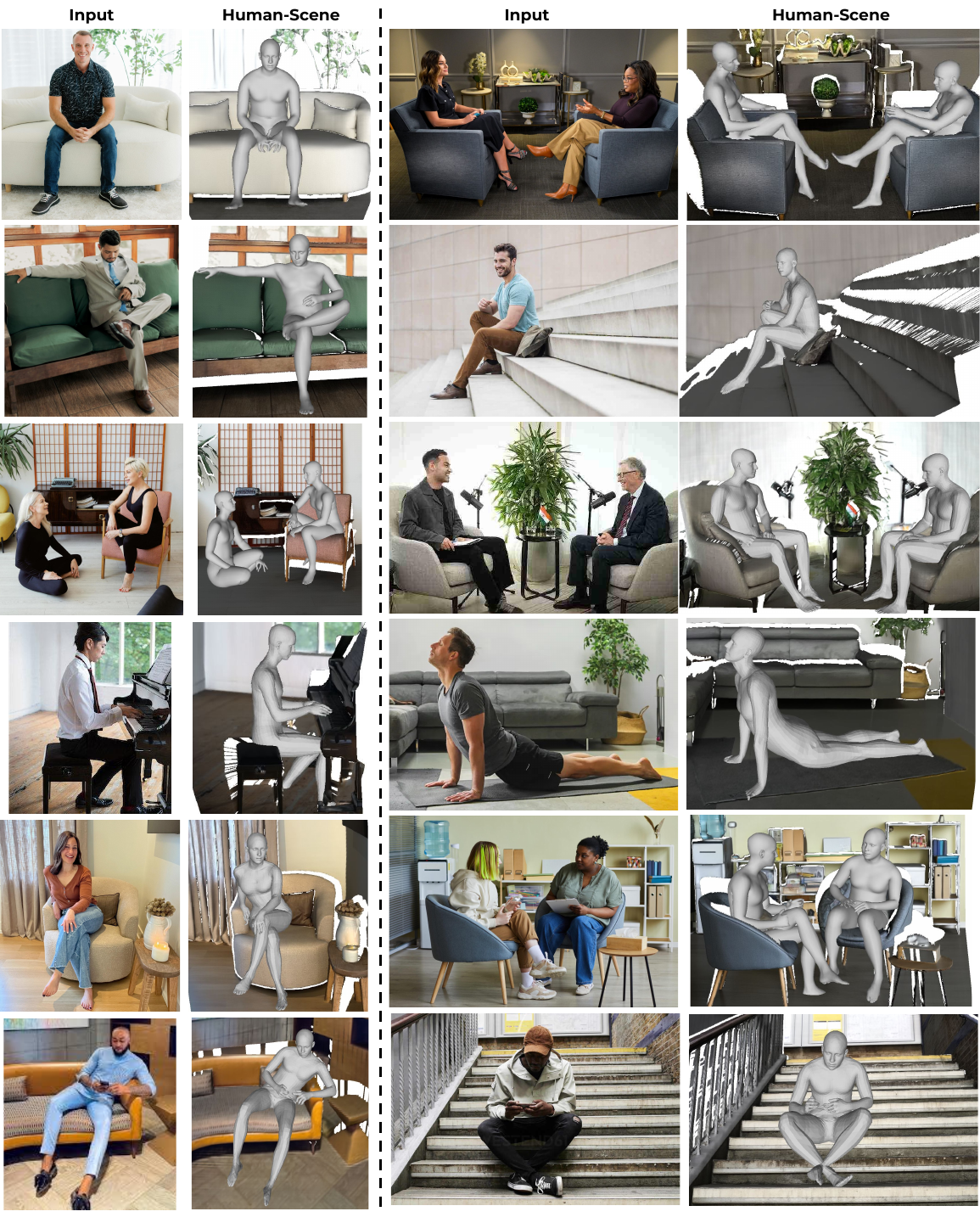} %
    \caption{\textbf{Additional Qualitative results for in-the-wild images.} Please refer to video for animated results.} 
    \label{fig:qualitative_additional}
\end{figure*}

We provide additional qualitative results on internet images in Fig.~\ref{fig:qualitative_additional}. %

\section{Optimization Hyperparameters}
\noindent\newtext{\textbf{Loss Weights.}} For the initial metric-scale human mesh alignment against loss $L_{align}$ \newtext{(Eq.~4)}, we set $\lambda_{j2d}$ and $\lambda_{d}$ to 5000 and 0.6 respectively. In the final joint optimization with respect to $L_{total}$ \newtext{(Eq.~6)}, we use the following loss weights: 

\begin{align*}
    \lambda_{j2d} &= 10000 \\
    \lambda_{d} &= 0.005 \\
    \lambda_{c} &= 0.7 \\
    \lambda_{i} &= 0.3 \\
    \lambda_{reg} &= 45
\end{align*}

\noindent\newtext{We ensure stability by applying all losses (except $L_{j2d}$) in the common metric vertex space, while $L_{j2d}$ is applied in normalized 2D coordinates (Sec.~3.3), making its value independent of image resolution. In addition, we set the loss weights such that the different terms have roughly similar magnitudes, which we found effective across all datasets.}

\noindent\newtext{\textbf{Termination Criteria.} 
Loss magnitude depends on the type of HSI, making loss-value termination criteria problematic and runtimes variable. For instance, a standing pose has far fewer contact/interpenetration vertices than a sitting pose. Hence, we terminate optimization based on iteration count. We observed that optimization of Eq.~3 and Eq.~6 converges within $30$ and $70-100$ iterations respectively, in most cases. The L-BFGS optimization in Eq.~4 converges rapidly, requiring just two iterations. This is since optimization of translation with respect to $L_d$ and fixed $\set{P}_h$ is convex.}

\section{License for In-the-wild Images}
\newtext{To demonstrate PhySIC's generalizability, we include qualitative results on Internet-sourced images and videos. The data was collected from (i) publicly available YouTube videos, used under fair-use provisions for research purposes, and (ii) image repositories that provide permissive licenses (e.g., Pexels).}

\section{Failure Cases}
\begin{figure*}[h!]
    \centering
    \includegraphics[width=0.95\linewidth]{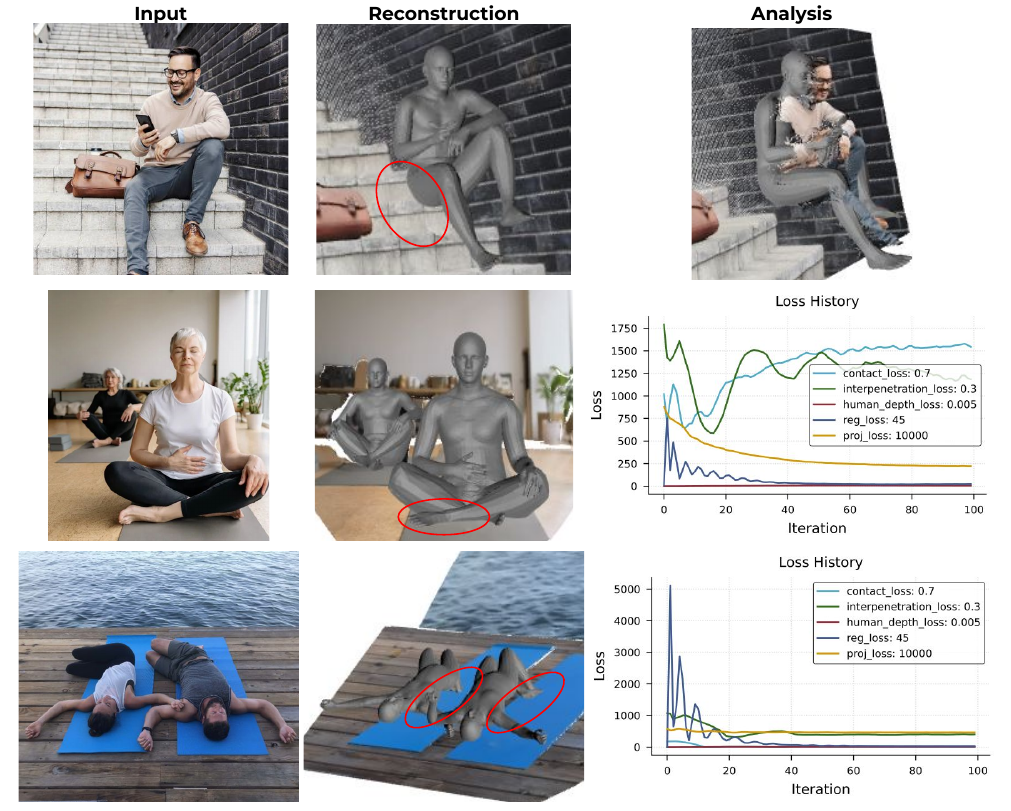} %
    \caption{\newtext{\textbf{Various Modes of Failure.} 
\textbf{Top:} Heavily floating and interpenetrating humans arise from misaligned human points $\mathcal{P}_h$, leading to incorrect localization within the scene. 
\textbf{Middle:} Despite explicit interpenetration losses, optimization over a multi-term objective and small scene geometry errors can still result in some penetration. 
\textbf{Bottom:} Missed contact predictions in initialization cause floating humans.}} 
    \label{fig:qualitative_failures}
\end{figure*}

\newtext{
While our approach is robust and generalizes to in-the-wild images, it can fail in certain cases. Our metric human point alignment (Sec. 3.2) assumes roughly similar scene geometries in $\hat{\set{P}_s}$ and $\set{P}_s$, an assumption that generally holds for robust depth models. In some cases, disagreeing predictions for $\set{I}$ and $\set{I}_s$ can cause inaccurate alignment despite RANSAC, resulting in heavily floating/interpenetrating humans due to incorrect localization (Fig.~\ref{fig:qualitative_failures}, row 1). Further, we highlight various edge cases:}

\noindent\newtext{\textbf{Floating Humans} Slight floating of the humans can be caused by false negatives in contact prediction by DECO. In such cases, using static contacts as discussed in Sec. G can improve the reconstruction.}

\noindent\newtext{\textbf{Unnatural Hands} In extreme cases, erroneous 2D hand poses and/or HSI terms can produce unnatural hand configurations. This can be addressed by incorporating joint-angle constraints or learned priors such as GRAB~\cite{taheri2020grab}.}

\noindent\newtext{\textbf{Unmodeled Scene Deformations} As illustrated in Fig.~3 (row 3), interpenetrations can be caused by unmodeled scene deformations, suggesting a promising future direction.}

\noindent\newtext{\textbf{Self-Penetrations}  Self-penetration is currently unpenalized, but can be readily incorporated within PhySIC (e.g. SMPLify-XMC self-contact loss).}

\section{Comparison with HSfM and Technical Differences}
\newtext{HSfM~\cite{mueller2024hsfm} reconstructs 3D human-scenes from uncalibrated multi-view images by integrating a human-joint reprojection loss within DUST3R’s~\cite{wang2024dust3r} global optimization. In contrast, PhySIC reconstructs arbitrary 3D human-scenes and their interactions from a single monocular image. This difference has several consequences. HSfM benefits from rich multi-view geometry constraints, achieving low PA-MPJPE even with only 2D joint supervision. In monocular settings, however, projection ambiguity arises: perfect 2D alignment does not always correspond to improved 3D pose, leading to higher baseline PA-MPJPE (Tab.~3). While stronger regularization could partially mitigate this, our ablation results (Tab.~3) indicate that 2D-joint supervision alone is insufficient in the monocular case.}

\newtext{To overcome this limitation, PhySIC introduces additional losses that compensate for the missing constraints, enforcing physically plausible human-scene interactions and producing realistic contacts even under occlusion. Beyond supervision, several technical choices further distinguish PhySIC from HSfM, including its optimization strategy, constraint formulation, and explicit handling of occlusions. Together, these design choices enable PhySIC to generalize beyond the multi-view setting and address the unique challenges of monocular human-scene reconstruction.
}

\section{Qualitative Ablations}
\newtext{Here, we present scene reconstructions for each ablation described in Sec.~4.5. As seen in \cref{fig:occlusion_ablation}, the human's lower half is occluded, leading to partial 2D keypoint detections. Direct initialization from off-the-shelf human and scene estimates exhibits significant misalignment. This misalignment persists when the human is optimized with $L_{j2d}$, since the loss is only a 2D joint reprojection objective and is not scene-aware. To address this misalignment, we introduce contact and interpenetration losses as additional physical constraints on the human pose. However, these losses operate against the nearest scene points as described in Sec.~3.3, which can produce incorrect optimization and degenerate outputs, as illustrated in~\cref{fig:occlusion_ablation}. Adding the depth loss ensures that the human is appropriately localized within the scene, which in turn enhances the effectiveness of the contact and interpenetration losses.  }

\newtext{
While the depth loss improves overall localization, occluded parts of the human, such as the lower half in this example, do not receive 2D supervision. Without 2D guidance, these unobserved parts can deviate significantly from the input image configuration. This occurs because the contact and interpenetration losses alone can modify them arbitrarily. To prevent this, we exclude occluded parts from the interpenetration loss and apply strong regularization, ensuring that these regions maintain plausible poses while the visible parts are correctly aligned with the scene. 
}

\newtext{
Lastly, the absence of a floor plane in occluded regions, caused by incomplete scene geometry from the depth estimators can cause incorrect optimization. Without complete floor geometry, the contact loss $L_c$ forces predicted contact vertices (e.g., feet) to ``snap'' onto the nearest available scene points, which are often furniture surfaces rather than the actual ground. This mechanically increases the number of positive contact predictions, boosting recall metrics, but creates many false-positive contacts that harm precision. Additionally, this misalignment forces unnatural body translations to satisfy these incorrect contact constraints, degrading MPJPE. By fitting and extending the floor plane (as shown in Fig.~\ref{fig:occlusion_ablation}), we provide correct contact targets for ground-contact vertices, which may reduce recall (fewer false positives) but improves both precision and pose accuracy by enabling physically plausible human-floor interactions.
}

\section{Robustness to Static Initial Contacts}
\newtext{
In addition to the ablations presented above, we test the robustness of our system to alternative initial contact definitions. Instead of using DECO-predicted contacts, we initialize optimization with a set of predefined static contacts from PROX~\cite{hassan2019prox}. These contacts correspond to body regions that are commonly in contact with the environment (e.g., back, bottom, and feet).
}

\newtext{
Results on the PROX-Quantitative dataset show that our method achieves similar performance to DECO-based initialization: a contact F1 score of 0.538 (static) versus 0.511 (DECO), and PA-MPJPE of 41.96 versus 41.99. This robustness can be attributed to the distance thresholding used in the contact and interpenetration losses, which mitigates false positives, as well as occlusion masking and related regularization strategies that stabilize optimization under incomplete 2D supervision. An implementation of this variant is included in our code repository.
}

